\documentclass[letterpaper]{article} 
\usepackage{aaai25}  
\usepackage{times}  
\usepackage{helvet}  
\usepackage{courier}  
\usepackage[hyphens]{url}  
\usepackage{graphicx} 
\usepackage{natbib}  
\usepackage{caption} 
\usepackage{algorithm}
\usepackage{algorithmic}
\usepackage{subfig}
\usepackage{amsmath}

\usepackage{newfloat}
\usepackage{listings}
\DeclareCaptionStyle{ruled}{labelfont=normalfont,labelsep=colon,strut=off} 
\floatstyle{ruled}
\newfloat{listing}{tb}{lst}{}
\floatname{listing}{Listing}
\title{ Multi-view Granular-ball Contrastive Clustering}
\author{
    Peng Su\textsuperscript{\rm 1,\rm 2},
    Shudong Huang\textsuperscript{\rm 1,\rm 2}\thanks{Corresponding author},
    Weihong Ma\textsuperscript{\rm 3},
    Deng Xiong\textsuperscript{\rm 4},
    Jiancheng Lv\textsuperscript{\rm 1,\rm 2} \\
}
\affiliations{
    \textsuperscript{\rm 1}College of Computer Science, Sichuan University, Chengdu 610065, China\\
    \textsuperscript{\rm 2}Engineering Research Center of Machine Learning and Industry Intelligence, Ministry of Education, China\\
    \textsuperscript{\rm 3}Information Technology Research Center, Beijing Academy of Agriculture and Forestry Sciences, Beijing 100097, China\\
    \textsuperscript{\rm 4}Stevens Institute of Technology, 1 Castle Point Terrace, Hoboken, NJ 07030, USA\\
    supeng@stu.scu.edu.cn, huangsd@scu.edu.cn, mawh@nercita.org.cn, dxiong@stevens.edu, lvjiancheng@scu.edu.cn
    
}

\usepackage{bibentry}

\begin{document}

\maketitle

\begin{abstract}
Previous multi-view contrastive learning methods typically operate at two scales: instance-level and cluster-level. 
Instance-level approaches construct positive and negative pairs based on sample correspondences, aiming to bring positive pairs closer and push negative pairs further apart in the latent space.
Cluster-level methods focus on calculating cluster assignments for samples under each view and maximize view consensus by reducing distribution discrepancies, e.g., minimizing KL divergence or maximizing mutual information. 
However, these two types of methods either introduce false negatives, leading to reduced model discriminability, 
or overlook local structures and cannot measure relationships between clusters across views explicitly.
To this end, we propose a method named Multi-view Granular-ball Contrastive Clustering (MGBCC). 
MGBCC segments the sample set into coarse-grained granular balls, and establishes associations between intra-view and cross-view granular balls. 
These associations are reinforced in a shared latent space, thereby achieving multi-granularity contrastive learning.
Granular balls lie between instances and clusters, naturally preserving the local topological structure of the sample set.
We conduct extensive experiments to validate the effectiveness of the proposed method. 

\end{abstract}

%

\section{Introduction}
Multi-view data refers to data collected from different sensors or obtained by different feature extractors, often exhibiting heterogeneity \cite{mvclustering_survey}.
For example, a web page typically contains images, text, and videos, each of which can be considered a view, reflecting the same sample from different perspectives. Multi-view clustering has received continuous attention in recent years, aiming to partition multi-view data into different clusters in an unsupervised manner \cite{awmvc_pr2019,mdgl_tkde2022,simple_mkkm_tpami2023,pimc_tnnls2024}. 
The key challenge in multi-view clustering is balancing the consistency and diversity between different views to learn the most comprehensive shared representation. 
Traditional multi-view learning methods mainly include subspace learning, graph learning, and multi-kernel learning. 
These methods often involve matrix decomposition and fusion, resulting in high computational complexity, making them difficult to apply to large-scale datasets, which severely hinders their practical application.

In recent years, deep-based multi-view learning methods have gained significant attention due to their excellent representation capabilities.
These methods extend deep single-view clustering and typically select appropriate feature extractors based on the properties of the views. 
DCCA \cite{dcca} projects data from two views into a common space using deep neural networks, where the representations of the two views are highly linearly correlated, making it a nonlinear extension of canonical correlation analysis. 
PARTY \cite{party} is a deep subspace clustering method with sparse priors that projects input data into a latent space, maintaining local structure by minimizing reconstruction loss while introducing sparse prior information into the latent representation learning to preserve sparse reconstruction relationships across the entire dataset. 
DIMC \cite{dimc} extracts high-level features of multiple views through view-specific autoencoders and introduces fusion graph-based constraints to preserve the local geometric structure of the data. 
DMCE \cite{dmce} applies ensemble clustering to fuse similarity graphs from different views, using graph autoencoders to learn a common spectral embedding. 
It designs a unified optimization framework to simultaneously minimize graph reconstruction loss, orthogonality loss, and graph contrastive loss. 
These methods combine deep learning with traditional multi-view learning ideas by introducing constraints from traditional methods such as neighbor graph constraints or self-expression constraints into the latent space projected by deep modules. 
This allows the models to learn concise but comprehensive representations that maximally preserve the structural information of the input data.

Multi-view contrastive learning is another important branch of deep multi-view clustering.
It can generally be divided into two categories: instance-level contrastive learning and cluster-level contrastive learning. 
The basic idea of the former is that instances of the same sample from different views should be as close as possible in the latent space, typically forming positive pairs. 
In these methods \cite{cmk, rcmk, dealmvc,sem}, the construction and handling of negative pairs is a key focus due to the unknown instance labels in unsupervised paradigms. 
Improper negative pairs can degrade the model's discriminative ability.
Cluster-level methods \cite{mflvc, cspan, cvcl}, on the other hand, align the clustering assignments of different views. 
They typically establish one-to-one correspondences between clusters across views and aim to make the distributions of corresponding clusters as consistent as possible. 
However, clusters are macro structures and do not effectively utilize the local structural information within views.

We propose a multi-granularity multi-view learning method. 
This method models the local structure of the sample set using granular balls and establishes intra-view and inter-view granular-ball connections based on overlap and intersection size, respectively. 
By bringing connected granular balls closer in the latent space, our model learns highly discriminative features.
To the best of our knowledge, this is the pioneering work utilizing granular-ball methodology for multiview contrastive learning. 
Specifically, our contributions are summarized as follows:
\begin{itemize}
    \item We propose a novel deep multi-view clustering method that performs contrastive learning at the granular-ball level. 
    This method avoids directly using neighboring samples to construct negative pairs while preserving the local structural information of the sample set, addressing the shortcomings of instance-level and cluster-level methods.
    \item We introduce a simple yet effective granular-ball construction method. Unlike classical methods that continuously bisect the dataset until reaching the smallest granularity, our method directly partitions the sample set into multiple granular balls based on the granularity parameter, avoiding the drawback of non-adjacent samples being grouped into the same granular ball in boundary regions.
    \item Extensive experiments on seven typical multi-view datasets demonstrate that our method achieves comparable or superior performance compared to state-of-the-art methods.
\end{itemize}

\section{Related Work}
In this section, we briefly review the latest advancements in related topics, including multi-view contrastive learning and granular-ball computing.

\subsection{Multi-view Contrastive Learning}
Contrastive learning \cite{moco, simclr, fnsgcl} aims to learn a feature space with good discriminative properties, where positive pairs are pulled closer together, and negative pairs are pushed further apart. 
In a single-view setting, positive pairs are typically constructed through augmentations of the same sample, while negative pairs come from other samples within the same batch or dynamically constructed feature queues. 
This idea naturally extends to multi-view learning, as multi-view instances can be seen as natural augmentations of a sample: they are unique yet collectively describe the same sample, giving rise to multi-view contrastive learning.

Completer \cite{completer} uses conditional entropy and mutual information to measure the differences and consensus between different views. By maximizing the mutual information between views, they aims to learn rich and consistent representations.
To resolve the conflict between maintaining multi-view semantic consistency and the reconstruction process that tends to preserve view-specific information, MFLVC \cite{mflvc} proposes a multi-level feature multi-view contrastive framework. 
This model learns low-level features, high-level features, and semantic labels from raw features in a fusion-free manner. 
Reconstruction is performed on low-level features, while consensus is explored through contrastive learning on high-level features.
SURE \cite{sure} addresses the issue of false negatives in multi-view contrastive learning, where two instances used to construct a negative pair might actually belong to the same cluster. 
It divides negative pairs into three intervals based on distance, treating those with distances below a certain threshold as potential positives for optimization.

\subsection{Granular-ball Computing}
\label{grcomputing}
Expanding on the theoretical foundations of traditional granularity computation and integrating the human cognitive mechanism of 'macro-first' \cite{ts_s1982}, Wang \cite{dgcc_sc2017} introduced the innovative concept of multi-granularity cognitive computation. Building on Wang's framework, Xia \cite{gb_is2019} developed an efficient, robust, and interpretable computational method known as granular-ball computing. Unlike traditional methods that process data at the most granular level of individual points, granular-ball computing encapsulates and represents data using granular-balls, thereby enhancing efficiency and robustness. 
Notable applications of granular-ball computing include granular-ball clustering \cite{fgb_tnnls2024, aef_icde2024, wgbc_icde2024, mgnr_tpami2024}, granular-ball classifiers \cite{gbsvm, gbtsvm_tcss2024}, granular-ball sampling methods \cite{gbsnlc_tnnls2023}, granular-ball rough sets \cite{gbnrs_tkde2022, gbrs_tnnls2024, ilbgb_tkde2023}, granular-ball three-way decisions \cite{3wc_gbnrs_tfs2024, 3wafgb_tfs2024}, and advancements such as granular-ball reinforcement learning \cite{gbrf}.

\begin{figure}[!htbp]
\centering
\includegraphics[scale=1.0]{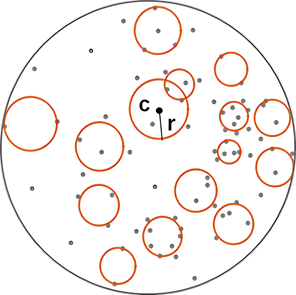}
\caption{Examples of granular balls}
\label{fig:gb_instance}
\end{figure}

Given a dataset $\{x_i\}_{i=1}^n$, 
let $\{GB_i\}_{i=1}^k$ denote the set of granular balls generated based on it, where $k$ represents the total number of balls.
As illustrated in Figure \ref{fig:gb_instance}, one ball contains multiple neighboring samples or feature points, e.g., $GB_i = \{ x_j\}_{j=1}^{n_i}$ , which essentially reflects the local topological relationships among samples. The center $c_i$ and the radius $r_i$ of $GB_i$ are defined as 
\begin{equation}
    \begin{gathered}
        c_i=\frac{1}{n_i}\sum_{j=1}^{n_i}x_j,\quad
        r_i=\frac{1}{n_i}\sum_{j=1}^{n_i}\|c_i-x_i\|_2.
    \end{gathered}
\end{equation}

In granular-ball computation, the key lies in how to generate the granular-ball set, which involves two critical steps: partitioning and merging. Partitioning refers to the recursive process of dividing a large ball into two smaller ones. 
Initially, the entire dataset is initialized as a single granular ball. 
Granular balls that meet the split conditions will continue to split. The split conditions typically vary depending on the task. In clustering tasks, if the average radius of the original ball is greater than the weighted average radius of the two sub-ball combined, the ball will split. Otherwise, it will stop. This condition can lead to over-partitioning, such as having one ball per sample. To prevent this, a minimum capacity threshold $\eta$ is introduced. If the number of samples in a ball is less than $\eta$, it will also stop splitting.

Merging refers to the process of combining two significantly overlapping ball into a single ball and recalculating the ball center and radius. Two balls are considered overlapping if they satisfy following conditions
\begin{equation}
    \label{merge_condition}
    \begin{gathered}
        \|c_i-c_j\|_2-(r_i+r_j)<\omega,\omega=\frac{min(r_i,r_j)}{min(p_i,p_j)}.
    \end{gathered}
\end{equation}
where $p_i$ and $p_j$ denote the total number of overlaps with adjacent granular balls for $GB_i$ and $GB_j$.
The merging process continues until the ball set no longer changes.

\begin{figure*}[!t]
\centering
\includegraphics[scale=0.5]{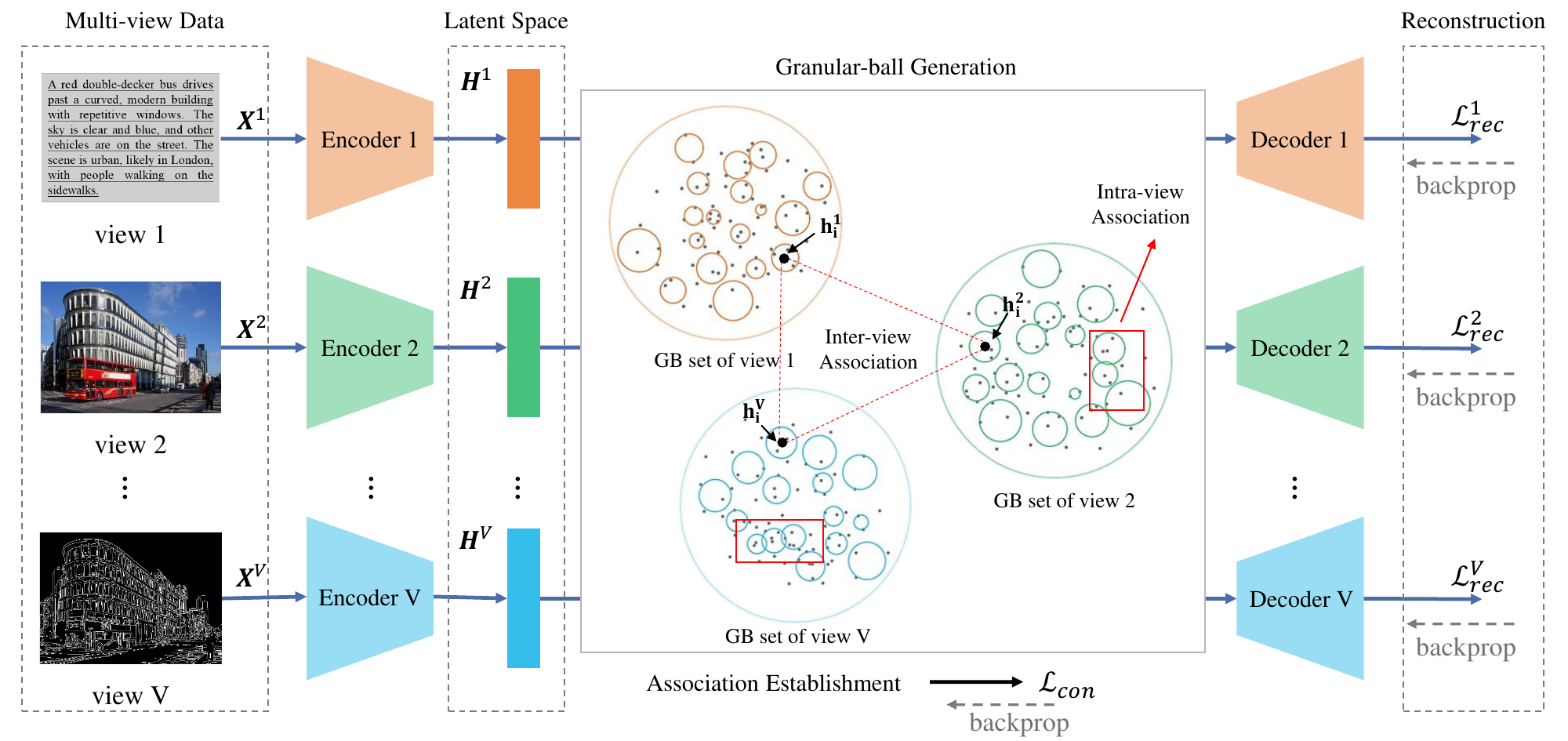}
\caption{
The framework of MGBCC. As shown, the overall loss function consists of two parts, e.g., reconstruction loss and granular-ball contrastive loss. 
We construct granular-ball sets $\{S^v\}_{v=1}^V$ for different views in the latent space and establish intra-view and cross-view associations based on overlap and intersection size respectively. 
Granular balls model the local structure of the dataset, and associated granular balls should be close to each other in the latent space. 
}
\label{fig:framework}
\end{figure*}

\section{Methodology}
In this section, we introduce a deep multi-view clustering method called Multi-view Granular-ball Contrastive Clustering (MGBCC). 
MGBCC encompasses four crucial processes: within-view reconstruction, within-view granular-ball generation, and cross-view granular-ball asociation and  granular-ball contrastive learning. The framework is shown in Figure \ref{fig:framework}.

\subsection{Within-view Reconstruction}
Given a multi-view dataset $\left\{\mathbf{X}^v\right\}_{v=1}^V$ with $N$ samples, each sample has instances from $V$ different views. Let $d_v$ represent the feature dimension of the $v$-th view, which typically varies across different views. To standardize the dimensions across views for subsequent comparison and fusion, we project the features of different views into a common dimension $d$. 

Deep autoencoders are employed as the representation learning framework to effectively extract essential low-dimensional embeddings from raw features.
We assign an autoencoder to each view. Specifically, for the $v$-th view, $E_v(\boldsymbol{\cdot};\theta^{v})$ and $D_v(\boldsymbol{\cdot};\phi^{v})$ denote its encoder and decoder, with $\theta^{v}$ and $\phi^{v}$ being their learnable parameters. As mentioned earlier, we set the output feature dimension of all encoders $\{E_v\}_{v=1}^V$ to $d$. After projection through encoders $\{E_v\}^V$, we obtain the high-level features $\left\{\mathbf{H}^v\right\}_{v=1}^V$ of instances from the $v$-th view by minimizing
\begin{equation}
    \begin{gathered}
        \mathcal{L}_{rec}=
        \sum_{v=1}^V\sum_{i=1}^N
        \Vert
        \mathbf{x}_i^v-D_v(E_v(\mathbf{x}_i^v;\theta^{v});\phi^{v})
        \Vert_2^2,
    \end{gathered}
\end{equation}
where $\mathbf{x}_i^v$ denotes the $i$-th sample of $\mathbf{X}^v$. The representation of $i$-th sample in $v$-th view is given by
\begin{equation}
    \begin{gathered}
        \mathbf{h}_i^v=E_v(\mathbf{x}_i^v;\theta^{v}).
    \end{gathered}
\end{equation}
In the subsequent computations, we will use these instance representations for constructing granular balls instead of the original features.

\subsection{Within-view Granular-ball Generation}
In clustering tasks, the classic granular ball partitioning method controls the granularity of division using a minimum capacity threshold (e.g., $\eta$). Granular balls are recursively split until the number of samples in a granular ball is less than this threshold. The issue with this method arises when it reaches the edge regions of the sample space, where the number of samples within a granular ball may be below the threshold, but the samples are dispersed (e.g., outliers). In such cases, the ball center may experience significant deviation, and the radius may be overestimated, resulting in inappropriate overlapping relationships.

To address this, we designed an alternative granular ball generation method. 
First, we introduce a granularity control parameter $p$, which roughly reflects the granularity of granular ball generation. 
Let $k$ denote the total number of granular balls to be generated. 
We set $N$ and $k$ to satisfy the following relationship
\begin{equation}
    \begin{gathered}
         k = \max\left(\left\lfloor \frac{N}{p} \right\rfloor, 1\right).
    \end{gathered}
\end{equation}

We then directly apply the $k$-means \cite{kmeans} to the entire dataset, dividing it into $k$ clusters, with each cluster considered a granular ball. For each ball, we compute its center and radius. Intuitively, outliers in the sample space, being far from most samples, tend to form their own clusters. Non-outliers will cluster together approximately according to the granularity parameter $p$. 

We construct granular balls for each view individually. 
For the $v$-th view, $H^v$ denotes latent representations obtained through the projection by the encoding layer $E^v$. Using the aforementioned method, we obtain 
$S^v=\{GB_i^v\}_{i=1}^{k_v}$, where $k_v$ represents the number of granular balls in the $v$-th view. 
If we set the granularity parameter $p_v$ to the same value $p$ for each view, the numbers $\{k_v\}_{v=1}^V$ across all views will be the same. 
For simplicity, we adopt this setting in the subsequent analysis and experiments.

Let $S=\{S^v\}_{v=1}^V$ denote granular ball sets constructed for all views, $\mathbf{C}^v$ be the center matrix of the $v$-th view, and $\mathbf{r}^v$ be the radius matrix. 
For a granular ball $GB_i^v$, its center is $\mathbf{c}_i^v$ and its radius is $r_i^v$. 
Note that the calculations of the centers and radii are gradient-preserving. 
In our granular-ball generation process, there is no merging process as in classical methods. 
However, we still need to consider the overlapping relationships of balls within each view.

Let $\mathbf{D}^v$ represent the center distance matrix of the $v$-th view. The distance between the $i$-th ball and the $j$-th ball is calculated as $\mathbf{d}_{ij}^v=\|\mathbf{c}_i^v-\mathbf{c}_j^v\|_2$. Based on $\mathbf{D}^v$, we compute the granular balls overlapping matrix $\mathbf{A}^v$ , which satisfies
\begin{equation}
    \begin{gathered}
        a_{ij}^v=
        \begin{cases}
            1, & \text{if Eq(\ref{merge_condition}) is satisfied}\\
            0, & \text{otherwise}
        \end{cases}.
    \end{gathered}
\end{equation}
Note that $\mathbf{A}^v$ will serve as part of the mask matrix for contrastive learning.

\subsection{Cross-view Granular-ball Association}
Through matrices $\{\mathbf{A}^v\}_{v=1}^V$, we have established the intra-view relationships between granular balls. 
Then we need to consider how to establish connections between cross-view granular balls. 
An intuitive approach is to consider two balls from different views as neighbors in the latent space if they each contain instances of the same sample from their respective views.
However, this method lacks robustness. 
When $p$ is relatively large, granular balls also become larger, and two cross-view balls might contain a very small number of common samples due to randomness.

To address this, we modified the method. Let $GB_i^{m}$ and $GB_j^{n}$ be two granular balls from views $m$ and $n$, respectively, containing $t_i$ and $t_j$ samples. First, we identify the common sample set between the two balls based on the stored sample indices:
\begin{equation}
    \begin{gathered}
        \text{Id}_{\text{both}} = \text{Id}(GB_i^{m}) \cap \text{Id}(GB_j^{n})
    \end{gathered}
\end{equation}
where $\text{Id}(\boldsymbol{\cdot})$ represents the sample indices contained in a granular ball. Next, we count the number of samples in $\text{Id}_{\text{both}}$:
\begin{equation}
    \begin{gathered}
        t_{\text{both}} = \text{length}(\text{Id}_{\text{both}})
    \end{gathered}
\end{equation}

Let $\mathbf{P}^{(m,n)}$ be the cross-view granular ball association matrix, which satisfies the following condition
\begin{equation}
    \begin{gathered}
        \text{p}_{ij}^{(m,n)} = 
        \begin{cases} 
            1, & \text{if } t_{\text{both}} / \min(t_i, t_j) \geq \tau \\
            0, & \text{otherwise}
        \end{cases}
    \end{gathered}
\end{equation}
Here, $\tau$ is a threshold parameter that determines the minimum proportion of common samples required to consider two cross-view granular balls as associated. 

\subsection{Granular-ball Contrastive Learning}
Matrices $\{\mathbf{A}^v\}_{v=1}^V$ reflects whether any two granular balls within a view overlap, while matrices $\{\mathbf{P}^{(m,n)}\}_{\forall{m \neq n}}$ indicates whether any two balls across views have sufficient intersection. We aim for these associated granular ball pairs to be as close as possible in the latent space, while unrelated pairs should be far apart. 

We use the granular-ball centers to represent the entire granular balls during the calculations. 
To facilitate computation, for any two views $m$ and $n$,  we define the combined center matrix as
\begin{equation}
    \begin{gathered}
        \mathbf{C} = 
        \begin{bmatrix} 
            \mathbf{C}^{m} \\ 
            \mathbf{C}^{n} 
        \end{bmatrix} 
    \end{gathered}
\end{equation}
then concatenate $\{\mathbf{A}^{m},\mathbf{A}^{n}\}$ and 
$\{\mathbf{P}^{(m,n)}, \mathbf{P}^{(n,m)}\}$ into a unified mask matrix
\begin{equation}
    \begin{gathered}
        \mathbf{M} = \begin{bmatrix}
            \mathbf{A}^{m} & \mathbf{P}^{(m, n)} \\
            \mathbf{P}^{(n, m)} & \mathbf{A}^{n}
        \end{bmatrix}
    \end{gathered}
\end{equation}
where matrix $\mathbf{P}^{(n, m)}$ is the transpose of $\mathbf{P}^{(m, n)}$. This unified mask matrix $\mathbf{M}$ ensures that granular balls within the same view and across different views are appropriately considered in the contrastive learning process. we calculate the contrastive loss at the granular-ball level
\begin{equation}
    \begin{gathered}
        \mathcal{L}^{(m, n)}=\frac{1}{k} 
        \sum_i^k 
        \sum_{j \in \Omega_i}
        \frac{exp(cos(\mathbf{c}_i, \mathbf{c}_j))}
        {\sum_{z\in \Phi_i} exp(cos(\mathbf{c}_i, \mathbf{c}_z))} 
    \end{gathered}
\end{equation}
where $\Omega_i=\{j|\mathbf{M}_{ij}=1, \forall{j}\}$ and $\Phi_i=\{z|\mathbf{M}_{iz}=0, \forall{z}\}$. 
$k$ represents the total number of granular balls and
$cos(\boldsymbol{\cdot}, \boldsymbol{\cdot})$ denotes the cosine similarity between two vectors. 
\begin{equation}
    \begin{gathered}
        \mathcal{L}_{con}=\frac{2}{V(V-1)}\sum_{\forall m \neq n} \mathcal{L}^{(m, n)}
    \end{gathered}
\end{equation}
We perform the same calculation process between any two views and take the average of all the losses as the final contrastive loss.
\subsection{Overall Loss And Optimization}
Combining the above two loss functions with a regularization parameter $\lambda$, the overall loss is formulated as
\begin{equation}
    \begin{gathered}
        \mathcal{L} = \mathcal{L}_{con} + \lambda \mathcal{L}_{rec}.
    \end{gathered}
\end{equation}
Any gradient-based optimization method can be used to minimize this objective function. 
We will further discuss the implementation details later.

\begin{table}
\caption{Description of the seven multi-view datasets.}
\centering
\renewcommand{\arraystretch}{1.2}
\setlength{\tabcolsep}{2pt} 
\begin{tabular}{ccccc}
\hline
\textbf{Dataset}       & \textbf{Samples} & \textbf{Clusters} & \textbf{Views} & \textbf{Dimensionality}         \\ \hline
BBCSport      & 544     & 5        & 2     & 3183/3203              \\
Caltech101-20 & 2386    & 20       & 2     & 1984/512 \\
Cora          & 2708    & 7        & 2     & 2708/1433              \\
Scene-15      & 4485    & 15       & 3     & 20/59/40               \\
MNIST-USPS    & 5000    & 10       & 2     & 784/784                \\
ALOI-100      & 10800   & 100      & 4     & 77/13/64/125           \\
NoisyMNIST    & 50000   & 10       & 2     & 784/784                \\ \hline
\end{tabular}

\label{dataset_details}
\end{table}

\begin{table*}[!t]
\caption{The clustering results on seven datasets (\%). The best results are bolded, and the second-best results are underlined. Results marked with a dot are directly quoted from the original papers. - indicates unavailable results due to out of memory.}
\centering
\renewcommand{\arraystretch}{1.2}
\begin{tabular}{ccccccccccc}
\cline{1-9}
\textbf{Dataset \textbackslash Method} & Completer   & MFLVC                & DealMVC        & DMCE  & CSPAN & ADPAC          & SURE           & \textbf{MGBCC}  &  &  \\ \cline{1-9}
\multicolumn{9}{c}{ACC(\%)}                                                                                                                                     &  &  \\ \cline{1-9}
BBCSport                               & 35.11       & 60.11                & {\underline{80.70$\boldsymbol{\cdot}$}}   & 37.13 & 58.27 & 35.29          & 55.33          & \textbf{95.77} &  &  \\
Caltech101-20                          & {71.42} & 36.92                & 40.36          & 61.48 & 44.72 & \textbf{77.16} & 50.21          & \underline{72.63}          &  &  \\
Cora                                   & 22.12       & 41.84                & \underline{49.07$\boldsymbol{\cdot}$}         & 26.48 & 42.54 & 30.87          & 42.76          & \textbf{65.44} &  &  \\
Scene-15                               & 39.29       & 32.87                & 32.71          & 34.85 & 33.73 & 41.49          & \underline{42.01}          & \textbf{43.72} &  &  \\
MNIST-USPS                             & 67.02       & 99.50$\boldsymbol{\cdot}$               & 80.92          & 62.18 & 80.14 & 98.16          & {\underline{99.56}}    & \textbf{99.60} &  &  \\
ALOI-100                               & 62.70       & 47.26                & 15.80          & 75.39 & 14.86 & 26.24          & {\textbf{90.37}}    & {\underline{88.39}}    &  &  \\
NoisyMNIST                             & 87.45       & 98.91                & \textbf{99.42} & -     & 55.68 & 97.27          & {\underline{99.14}}    & 98.48          &  &  \\ \cline{1-9}
\multicolumn{9}{c}{NMI(\%)}                                                                                                                                     &  &  \\ \cline{1-9}
BBCSport                               & 2.62        & 43.04                & {\underline{65.59$\boldsymbol{\cdot}$}}   & 4.39  & 48.27 & 3.94           & 38.78          & \textbf{87.21} &  &  \\
Caltech101-20                          & {70.96} & 52.68                & 59.41          & 57.93 & 63.04 & \textbf{73.85} & 60.92          & \underline{71.57}          &  &  \\
Cora                                   & 1.42        & 30.41                & \underline{37.75$\boldsymbol{\cdot}$}         & 1.19  & 22.21 & 7.65           & 22.49          & \textbf{46.86} &  &  \\
Scene-15                               & {\underline{43.46}} & 33.93                & 32.46          & 32.09 & 31.23 & \textbf{44.21} & 42.87          & 41.62          &  &  \\
MNIST-USPS                             & 81.83       & 98.50$\boldsymbol{\cdot}$               & 90.24          & 72.90 & 78.27 & 95.61          & {\underline{98.68}}    & \textbf{98.96} &  &  \\
ALOI-100                               & 84.00       & 75.72                & 56.12          & 83.67 & 39.88 & 51.59          & {\underline{94.04}}    & \textbf{94.78} &  &  \\
NoisyMNIST                             & 86.70       & 96.79                & \textbf{98.15} & -     & 60.07 & 93.34          & {\underline{97.30}}    & 96.17          &  &  \\ \cline{1-9}
\multicolumn{9}{c}{PUR(\%)}                                                                                                                                     &  &  \\ \cline{1-9}
BBCSport                               & 35.85       & 69.12                & {\underline{80.70$\boldsymbol{\cdot}$}}   & 37.13 & 69.30 & 38.62          & 63.60          & \textbf{95.77} &  &  \\
Caltech101-20                          & 78.58       & 69.74                & 70.12          & 71.12 & 76.19 & \underline{80.68} & 73.60          & {\textbf{81.89}}    &  &  \\
Cora                                   & 30.28       & 51.51                & {\underline{60.67$\boldsymbol{\cdot}$}}   & 30.32 & 49.89 & 36.74          & 48.15          & \textbf{68.02} &  &  \\
Scene-15                               & 42.88       & 34.02                & 34.34          & 36.77 & 36.39 & {\underline{45.08}}    & 44.93          & \textbf{47.07} &  &  \\
MNIST-USPS                             & 73.00       & 99.50$\boldsymbol{\cdot}$               & 80.92          & 66.80 & 81.70 & 98.16          & {\underline{99.56}}    & \textbf{99.60} &  &  \\
ALOI-100                               & 66.74       & 48.58                & 15.80          & 76.97 & 18.53 & 30.74          & \textbf{90.67} & {\underline{89.31}}    &  &  \\
NoisyMNIST                             & 87.45       & 98.91                & \textbf{99.42} & -     & 60.69 & 97.27          & {\underline{99.14}}    & 98.48          &  &  \\ \cline{1-9}
\end{tabular}
\label{tab:cluster_metrics}
\end{table*}

\section{Experiments}
In this section, we analyze the experimental results of the proposed method on seven widely used multi-view datasets and compare it with several state-of-the-art methods to demonstrate its effectiveness. 
\subsection{Experimental Settings}
\textbf{Datasets.} Seven multi-view benchmark datasets are employed in this work. 
BBCSport \cite{greene06icml} includes 544 sports news articles in 5 subject areas, with 3183-dimensional MTX features and 3203-dimensional TERMS features, forming 2 views. 
Caltech101-20 \cite{Li_Nie_Huang_Huang_2022} contains 101 classes in total. We select 20 widely used classes with 2 views and 2386 samples for our experiments.
Cora \cite{6413846} contains 4 views, including content, inbound, outbound, and citations, extracted from the documents. 
Scene-15 \cite{Fei-Fei_Li_Perona_2005} consists of 4568 natural scenes categorized into 15 groups. Each scene is described by three types of features: GIST, SIFT, and LBP.
MNIST-USPS \cite{Peng_Huang_Lv_Zhu_Zhou_2019} is a popular handwritten digit dataset containing 5000 samples with two different styles of digital images.
ALOI-100 \cite{schubert2010elki} consists of 10800 object images, with each image described by 4 different features.
NoisyMNIST \cite{Wang_Arora_Livescu_Bilmes_2015} uses the original images as view 1 and randomly selects within-class images with white Gaussian noise as view 2. 
Table \ref{dataset_details} lists the important information of all datasets.

\textbf{Compared Methods.}
We compared the proposed method with seven classical or state-of-the-art methods including Completer \cite{completer}, MFLVC \cite{mflvc}, DealMVC \cite{dealmvc}, DMCE \cite{dmce}, CSPAN \cite{cspan}, ADPAC \cite{adpac}, SURE \cite{sure}.
All compared methods are implemented according to the source codes released by the authors, and the hyper parameters are set according to the suggestion in the corresponding paper.

\textbf{Evaluation Metrics.}
To perform a fair comparison, we adopt the commonly used metrics, e.g., clustering accuracy (ACC), normalized mutual information (NMI), and purity (PUR).

\subsection{Implementation Details}
The network structure follows a standard autoencoder architecture. 
For each view, the encoder consists of several linear layers with ReLU activation functions between each pair of layers. 
Except for the BBCSport and Cora datasets, all other datasets use the same encoder structure with dimensions set as $\{d_v, 2000, 500, 500, d\}$, where $d_v$ is the input feature dimension of each view.
$d$ is the projection feature dimension, which is the same for all views. 
After encoding, inputs undergo standardization. 
The decoder mirrors the encoder structure. 
For the Cora dataset, we use the same dimensions but without activation functions between layers, resulting in a linear projection. 
For BBCSport, given its small sample size of 544, we use a single-layer linear projection with the encoder dimensions set to $\{d_v, d\}$.
 
Our implementation of MGBCC is carried out using PyTorch 2.3 \cite{pytorch} on a Windows 10 operating system, powered by an NVIDIA GeForce GTX 1660 Ti GPU. 
We employ the Adam optimizer with learning rate of 0.0001 and weight decay of 0. The batch size is typically set to either 256 or 1024, depending on the dataset size. 
The regularization parameter $\lambda$ is generally set to 1 across most datasets, except for BBCSport and Cora, where it is adjusted to 0 due to differences in the projection approach (e.g., linear or nonlinear). 
The threshold parameter $\tau$ is uniformly set to 0.1 across all datasets.
The granularity parameter $p$ significantly impacts the experimental results, which will be analyzed later.

During the clustering phase, we equally weight and fuse the projected features $\left\{\mathbf{H}^v\right\}_{v=1}^V$ from each view and then apply the $k$-means algorithm to obtain clustering labels.

\subsection{Experimental Results}
\begin{figure*}[!htbp]
\centering
\captionsetup[subfloat]{font=small}
\subfloat[Initial (ACC=58.40\%)]{\includegraphics[scale=0.75]{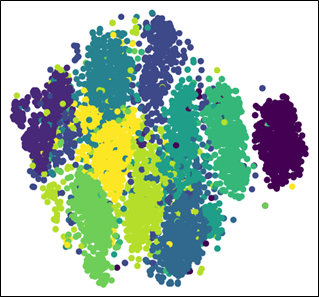}} \hspace{0.3cm}
\subfloat[1st epoch (ACC=74.42\%)]{\includegraphics[scale=0.75]{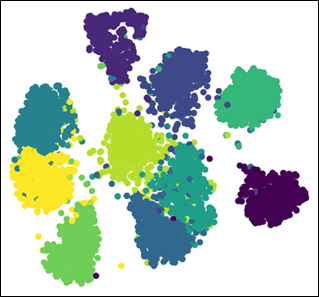}} \hspace{0.3cm}
\subfloat[5th epoch (ACC=98.20\%)]{\includegraphics[scale=0.75]{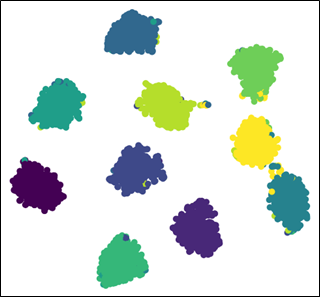}} \hspace{0.3cm}
\subfloat[20th epoch (ACC=99.54\%)]{\includegraphics[scale=0.75]{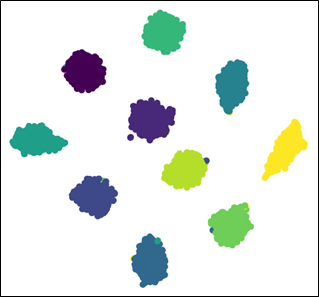}}
\caption{The t-SNE visualization of the clustering results on MNIST-USPS dataset.} 
\label{mnist_usps_cluster}
\end{figure*}

In Table \ref{tab:cluster_metrics}, we present the experimental results of the proposed method in comparison with other methods, leading to the following conclusions: 
(1) Across the three given metrics, the proposed method achieves the best or second-best results for most case. 
Even on the NoisyMNIST datasets, the proposed method ranks approximately third, with only slight differences from the top two methods.
Using Cora dataset as an example, our proposed method achieves an accuracy of 65.44\%, significantly surpassing the best comparative result of 49.07\%. 
This demonstrates the effectiveness and competitiveness of our method.
(2) Compared with classical multi-view constrastive learning methods (e.g., Completer, DealMVC, SURE), the proposed method consistently achieves more favorable clustering results across the majority of datasets. 
As a representative, SURE is an instance-level contrastive learning method that focuses on the issue of false negatives, achieving the best or second-best results on the Scene-15, MNIST-USPS, ALOI-100, and NoisyMNIST datasets. 
It’s noteworthy that the proposed method does not show significant performance degradation on these four datasets and clearly outperforms other methods. 
Moreover, on the remaining three datasets, the proposed method significantly outperforms SURE, highlighting the effectiveness of contrastive learning at the granular-ball level.
(3) We visualize the clustering results of the proposed method on the MNIST-USPS dataset using t-SNE. 
As shown in Figure \ref{mnist_usps_cluster}, the clustering structure becomes progressively clearer with the increase of the optimization epoch.
This suggests that the proposed method is effective at revealing the underlying cluster structure.
 
     

\subsection{Ablation Studies}
To accurately validate the effectiveness of contrastive learning at the granular-ball level, we conduct an ablation study on the Caltech101-20 dataset. 
The feature dimension \(d\) is set to 128. 
Based on this, we adopted three experimental settings. 
The first setting trains the model solely based on reconstruction loss. 
The second setting includes both the reconstruction loss and instance-level contrastive loss (i.e., $p=1$). 
The third setting incorporates reconstruction loss and granular-ball contrastive loss, with the granularity parameter set to 2. 
It is important to emphasize that the parameter $p$ reflects the average granularity rather than the absolute granularity.


\begin{table}
\centering
\caption{Ablation studies on Caltech101-20.}
\label{ablation_studies}
\renewcommand{\arraystretch}{1.2}
\setlength{\tabcolsep}{4pt} 
\begin{tabular}{lccc}
\hline
Setting & ACC (\%)       & NMI (\%)       & PUR (\%)       \\ \hline
$\mathcal{L}_{rec}$      & 40.28          & 60.03          & 75.40          \\
$\mathcal{L}_{rec} + \mathcal{L}_{con}\sim p=1$      & 44.47          & 62.38          & 78.54          \\
$\mathcal{L}_{rec} + \mathcal{L}_{con}\sim p=2$      & \textbf{72.63} & \textbf{71.57} & \textbf{81.89} \\ \hline
\end{tabular}
\end{table}

Table \ref{ablation_studies} presents the corresponding experimental results. 
As can be seen, the instance-level contrastive method performed poorly on this dataset, whereas the granular-ball contrastive method achieved significant improvements. 
This further demonstrates the feasibility and effectiveness of contrastive learning at the granular-ball level.

\subsection{Parameter Analysis }
The model has two important hyperparameters: the granularity parameter $p$ and the dimension $d$ of the projection features. 
The former essentially reflects the average size of the granular balls. 
When $p$ is set to 1, it is equivalent to instance-level contrastive learning. 
The latter affects the amount of original feature information contained in the latent representation. 
If $d$ is too small, important information might be lost, whereas if $d$ is too large, it increases the complexity of optimization and memory requirements.
To explore the optimal parameter settings, we conduct experiments on Caltech101-20 and Cora. 
We varied the parameter $p$ within the range [1, 2, 4, 8, 16], and the parameter $d$ within the range [8, 16, 32, 64, 128, 256]. 

\begin{figure}[!htbp]
\centering
\captionsetup[subfloat]{font=small}
\subfloat[Caltech101-20]{\includegraphics[scale=0.7]{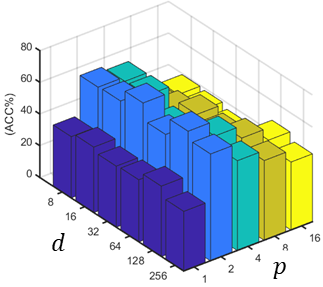}}
\hspace{0.3cm}
\subfloat[Cora]{\includegraphics[scale=0.7]{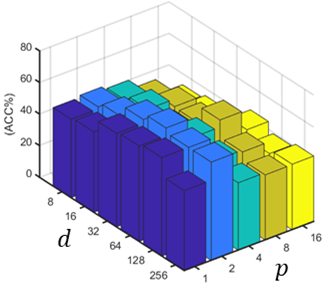}}
\caption{The clustering accuracy (\%) with different parameters $p$ and $d$ on Caltech101-20 and Cora.} 
\label{para_analysis}
\end{figure}

Figure \ref{para_analysis} illustrates the experimental results on the aforementioned datasets.
When $p$ is set to 2, the proposed method performs well. 
However, as $p$ increases, performance gradually declines. 
This may be because larger granular balls can no longer be effectively associated based solely on overlap and intersection size. 
When $p$ becomes too large, the method essentially degrades into a cluster-level contrastive approach, where reducing cluster assignment differences might be a more reasonable strategy.
In experiments, we typically set the parameter $p$ to 1, 2, or 4.
The parameter $d$ has a relatively minor impact on the experimental results and is generally set to 64.


\begin{figure}[!htbp]
\centering
\includegraphics[scale=1.0]{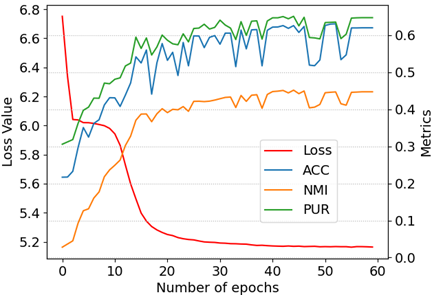}
\caption{Loss vs. Metrics on Cora.}
\label{convergence_analyze}
\end{figure}

\subsection{Convergence Analysis}
We evaluated the convergence of the proposed method on Cora dataset by tracking the loss values and corresponding clustering performance over increasing epochs. As shown in Figure \ref{convergence_analyze}, the total loss values gradually decrease and converge within 100 epochs. 
These results demonstrate the strong convergence performance of the proposed method.

\section{Conclusion}
In this paper, we propose a multi-view granular-ball contrastive clustering method. 
Specifically, we model the local structure of the sample set using granular balls in the latent space, resulting in respective granular-ball sets for each view. 
We establish intra-view and cross-view associations between granular balls based on their overlap and intersection size, encouraging associated granular balls to be close to each other in the latent space. 
Extensive experiments have been conducted to validate the effectiveness of the proposed method.
In the future, we will extend the proposed method to handle incomplete multi-view data.

\bibliography{aaai25}

\end{document}